\pdfoutput=1

\documentclass[11pt]{article}

\usepackage{graphicx}
\usepackage{algorithm}
\usepackage{algorithmic}
\usepackage{amsmath}
\usepackage{amsthm}
\usepackage{amssymb}
\usepackage{multirow}
\usepackage{booktabs}
\usepackage{longtable}
\usepackage[figuresright]{rotating}

\usepackage{color}
\usepackage{xcolor}
\definecolor{shadecolor}{rgb}{0.92,0.92,0.92}  
\usepackage{framed}

\usepackage{tcolorbox}
\usepackage{xcolor}
\newtcolorbox{myshaded}{
  colframe=black,
  colback=gray!7,
  boxrule=0.8pt,
  left=5pt, 
  right=8pt, 
  top=8pt, 
  bottom=8pt, 
}

\usepackage[]{acl}

\usepackage{times}
\usepackage{latexsym}

\usepackage[T1]{fontenc}

\usepackage[utf8]{inputenc}

\usepackage{microtype}

\usepackage{inconsolata}

\title{Good Idea or Not, Representation of LLM Could Tell}

\author{
Yi Xu\textsuperscript{1},\ 
Bo Xue\textsuperscript{1},\ 
Shuqian Sheng\textsuperscript{1},\ 
Cheng Deng\textsuperscript{1},\
Jiaxin Ding\textsuperscript{1}\\
\textbf{Zanwei Shen\textsuperscript{1}},\
\textbf{Luoyi Fu\textsuperscript{1}},\
\textbf{Xinbing Wang\textsuperscript{1}},\ 
\textbf{Chenghu Zhou\textsuperscript{2}} \\
        \textsuperscript{1}Shanghai Jiao Tong University, Shanghai, China \\
        \textsuperscript{2}IGSNRR, Chinese Academy of Sciences, Beijing, China\\
        \texttt{ yixu98@sjtu.edu.cn }
}

\begin{document}
\maketitle
\begin{abstract}
In the ever-expanding landscape of academic research, the proliferation of ideas presents a significant challenge for researchers: discerning valuable ideas from the less impactful ones. The ability to efficiently evaluate the potential of these ideas is crucial for the advancement of science and paper review. In this work, we focus on idea assessment, which aims to leverage the knowledge of large language models to assess the merit of scientific ideas. First, we investigate existing text evaluation research and define the problem of quantitative evaluation of ideas. Second, we curate and release a benchmark dataset from nearly four thousand manuscript papers with full texts, meticulously designed to train and evaluate the performance of different approaches to this task. Third, we establish a framework for quantifying the value of ideas by employing representations in a specific layer of large language models. Experimental results show that the scores predicted by our method are relatively consistent with those of humans. Our findings suggest that the representations of large language models hold more potential in quantifying the value of ideas than their generative outputs, demonstrating a promising avenue for automating the idea assessment process.
\end{abstract}

\section{Introduction}


The rapid pace of scientific research in various disciplines has given rise to an overwhelming number of academic papers~\cite{tabah1999literature, bornmann2015growth,xu-etal-2023-exploring}. Typically, ideas are commonly conveyed through these papers, which reviewers must carefully scrutinize to grasp the ideas authors present. However, amidst the vast volume of paper submissions, the review process becomes slow, labor-intensive, and less precise~\cite{checco2021ai}, making it a challenge to identify valuable ideas. Fortunately, with the advent of large language models (LLMs), we are presented with an unprecedented opportunity to revolutionize how we assess the merit of scientific ideas, and this work explores their use as a knowledge tool for such evaluation. In order to make idea more concrete, \textbf{we take unpublished manuscripts or the latest papers as the research object.}

While the generative capabilities of LLMs have been widely recognized, their potential as idea (or paper) evaluative instruments has remained relatively underexplored. Recent studies~\cite{yuan2022can,liang2023can,liu2023reviewergpt,agarwal2024litllm} have begun to harness LLMs for the automatic generation of paper reviews, aiming for outputs that are both informative and emulate human feedback. These efforts primarily utilize the text generation capability of LLM, which are highly dependent on the scale of model parameters. For instance, \citet{liu2023reviewergpt} proves that GPT-4 surpasses other open-source LLMs, such as LLaMA~\cite{touvron2023llama}, in generating reviews. Meanwhile, crafting intricate prompts containing specific commands and inquiries is essential for LLMs to produce meaningful reviews. Nonetheless, LLM-generated reviews can still reflect models' subjective biases and occasionally produce hallucinated contents~\cite{zhang2023siren,manakul-etal-2023-selfcheckgpt}. There is currently no research that quantitatively evaluate ideas in an objective way.

According to \citet{geva2021transformer,zou2023representation}, the representations of different layers in LLM contain different semantic information, and in some tasks, the performance of the last layer is not the best. Based on this point, we suppose that the representations of LLMs encapsulate a detailed and nuanced comprehension of text, which can be leveraged to construct a systematic and objective framework for the assessment of ideas. Our research thus focuses on the \textbf{quantitative evaluation of ideas} through LLMs, an approach we argue is more objective than generative techniques. It is worth noting that LLMs' generative method is not inherently adept at processing numerical data (digits), which is also why we turn to their representations as a means to quantify the value of ideas. We delve into the latent knowledge embedded within the representations of LLMs for this purpose. Specifically, we first define the problem of quantitative idea evaluation, and construct a benchmark dataset comprised of nearly 4k scientific manuscripts with full texts in the discipline of computer science. This dataset serves as the bedrock for training and testing various LLM-based approaches to idea assessment. In the subsequent phase of our research, we develop a framework for quantifying the value of ideas. This framework leverages the representations produced by LLMs, which encode semantic features of the text in a high-dimensional space. Finally, by training a downstream evaluator using these representations, our framework can identify patterns and signals within these representations that align with the value of scientific ideas as determined by the distribution of expert consensus.

We perform extensive experiments on our benchmark dataset. The results reveal that the representations generated by LLMs are inherently indicative of the potential value of scientific ideas, especially using the representations in the middle and rear layers, which demonstrate a high degree of consistency with human judgements. Meanwhile, we also found that high consistency can be achieved with only a small amount of data, thanks to the pretrained knowledge in LLMs. The contributions of our paper are summarized as follows:
\begin{itemize}
    \item We conduct a thorough investigation of current methods for paper (idea) assessment and pioneer the study of quantitative evaluation of scientific ideas.
    \item We meticulously curate a new benchmark dataset, complete with human-assigned scores on overall quality, novelty, and correctness. We are making it publicly available. This dataset is comprised of the full texts of 3,795 papers in PDF format.
    \item We propose a new framework that quantifies the value of ideas by utilizing the deep representations produced by LLMs.
    \item We conduct extensive experiments to verify the performance of our method. The results indicate that utilizing the representations from LLMs aligns more closely with human judgments than using generative textual outputs of LLMs when assessing the quality of ideas.
\end{itemize}

\section{Related Work}

\citet{yuan2022can} have explored the use of various NLP techniques to produce decisive and comprehensive reviews of academic papers from multiple perspectives. The work of ReviewerGPT~\cite{liu2023reviewergpt} studies the application of LLMs as review assistants, focusing on tasks such as error identification, checklist verification, and comparative paper analysis. Another innovative approach by researchers~\cite{liang2023can} involves an automated system that employs GPT-4 to create review comments and suggestions for revisions, which are then benchmarked against feedback from human reviewers. What's more, LitLLM~\cite{agarwal2024litllm} equips Retrieval Augmented Generation (RAG) module to address the hallucination problem in review generation.

In addition to the methods mentioned above, some researches rely on external data to assess the quality of a paper. For instance, \citet{thelwall2023predicting} have developed a framework that predicts article quality scores using a range of bibliometric and metadata indicators, including citation counts, journal impact factors, and institutional rankings. Similarly, KQI~\cite{wang2023quantifying} leverages the structure of citation networks to quantify the knowledge contribution of a paper. This kind of approaches, however, pertains to post-publication evaluation. Unlike our approach, which is based solely on the text of the paper itself, it does not require information about the paper's acceptance or publication status. 


\section{Idea Assessment}
In this work, we focus on quantitative evaluation of ideas. Let $D = \{d_1, d_2, ..., d_n\}$ be a dataset consisting of $n$ scientific manuscripts (papers), each representing a distinct scientific idea. The direct scoring of an idea involves mapping each manuscript $d_i$ to a quantitative score $s_i$ based on a predefined criterion $c \in C$. The criterion set $C$ serves as the basis for the assessment and is essential for guiding the evaluation process. It can encompass various aspects of potential impacts of an idea, such as novelty, correctness, excitement, soundness, or alignment with current research trends. 

We first define a function $A_{c}: D \rightarrow \mathbb{R}$ such that for any manuscript $d_i \in D$ and criterion $c \in C$, $A_{c}(d_i)$ produces a scalar value $s_i \in \mathbb{R} $ which quantifies the value of the idea $d_i$ with respect to $c$. The quantitative evaluation of idea $d_i$ with respect to criterion $c$ can be expressed as:

\begin{equation}
    s_i = A_{c}(d_i),
\end{equation} 
where the function $A_{c}$ is the evaluator that assesses the idea based on the text of the manuscript $d_i$ and the specified criterion $c$. 

In our work, we utilize the representations of LLM to quantify the value of an idea. Let $M$ be an LLM that transforms textual data into a high-dimensional representation space. We define an encoding function $Rep: D \rightarrow \mathbb{R}^{l \times m}$ such that for any manuscript $d_i \in D$, there is a hidden representation $Rep(d_i) \in \mathbb{R}^{l \times m}$, where $l$ is the number of tokens in $d_i$ and $m$ is the dimension of the representation space in LLM $M$. Now, we revise the evaluator function $A_{c}: \mathbb{R}^{l \times m} \rightarrow \mathbb{R}$ that maps the representation to a scalar value $s_i$:

\begin{equation}
    s_i = A_{c}(Rep(d_i)).
\end{equation} 

The evaluator, $A_{c}$, is designed to be flexible and adaptable to different criteria and can be trained using annotated data that provide ground truth measures of the ideas' impact with respect to the chosen criterion. To the best of our knowledge, we are the first to quantify the value of an idea.

\section{Dataset}

To ensure that the idea evaluator is well-calibrated, the benchmark idea dataset $D$ should be representative of the scientific community and contain cutting-edge knowledge in academia. To this end, we have compiled a collection of 3,795 manuscripts that are available in PDF format from the International Conference on Learning Representations (ICLR) 2023. For the extraction of full texts from these PDFs, we employed GROBID~\cite{lopez2009grobid}, a sophisticated tool for parsing academic PDF documents. 

Additionally, the metadata of these papers includes comprehensive evaluation criteria from official reviewers, encompassing scores for overall quality, correctness, technical and empirical novelty, providing a rich ground truth for training and validation. It is possible that an idea is interesting but the paper score of a criterion such as correctness is low because there are flaws in the experiments. Therefore, in our work, we mainly investigate the criterion \textbf{\textit{overall quality}} and take it as the overall score of an idea. 

\begin{table}[!htb]
\centering
\small
\tabcolsep 0.105in
\renewcommand{\arraystretch}{1.2}
\begin{tabular}{lcc}
\toprule[1pt]
                           & \textbf{ICLR23-low-std} & \textbf{ICLR23-all} \\ \hline
\textbf{\# paper}          & 1901                    & 3795                \\ \hline
\textbf{overall quality}     & 5.52 $\pm$ \textbf{0.61}       & 5.41 $\pm$ 1.06              \\
\textbf{correctness}       & 3.09 $\pm$ 0.44         & 3.09 $\pm$ 0.49              \\
\textbf{technical novelty} & 2.59 $\pm$ 0.43         & 2.59 $\pm$ 0.48              \\
\textbf{empirical novelty} & 2.56 $\pm$ 0.41         & 2.56 $\pm$ 0.47              \\ \bottomrule[1pt]
\end{tabular}
\caption{Statistics of benchmarks.}
\label{table_statistics}
\end{table}

Considering the different consistencies of human-rated data, which can have impacts on different evaluation models, we choose papers with highly consistent human-rated scores from the original dataset \textbf{\textit{ICLR23-all}} and construct dataset \textbf{\textit{ICLR23-low-std}}, where the standard deviation (std) of overall quality scores for each paper is relatively lower. The statistics are listed in Table~\ref{table_statistics}. 


\begin{figure*}[htb]
    \centering
    \includegraphics[width=1.0\linewidth]{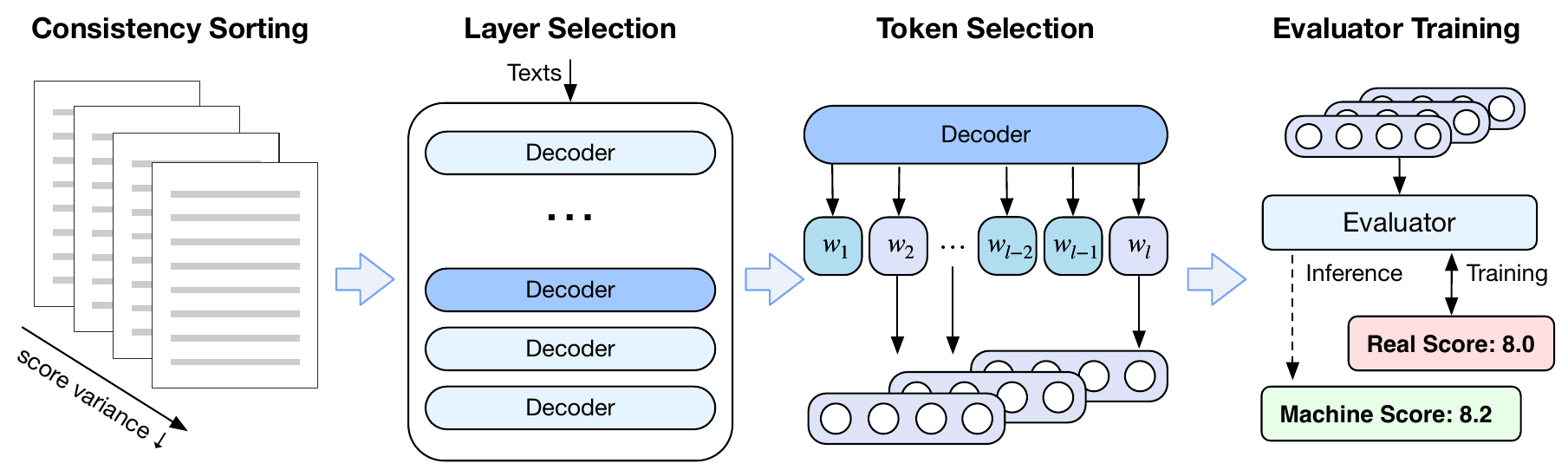}
    \caption{Pipeline of quantitative evaluation of ideas using the representations of LLMs.}
    \label{fig_ideval}
\end{figure*}

\section{Methodology}
The purpose of our method is to train an evaluator $A_c$ to score ideas, which consists of four steps: consistency sorting, layer selection, token selection, and evaluator training. Figure~\ref{fig_ideval} shows the pipeline of quantitative evaluation of ideas using the representation of LLMs. It should be highlighted that the steps of layer and token selection only exist in training process, which are determined during the inference process.

\subsection{Consistency Sorting}\label{c_sorting}
In our scenario, we anticipate that models can learn the rating standard from human-rated data. Specifically, the human-assigned scores for each paper in the training set should exhibit a high level of consistency; that is, the more uniform the scores for each paper are (reflected by lower variance), the more suitable the data is for model training. Therefore, our method employs a consistency-based sorting mechanism to construct the training and testing sets. We commence by ordering the papers according to the variance in human scores for a given criterion $c$. Subsequently, based on a predetermined threshold for training set partitioning, papers that demonstrate high consistency (low variance) are allocated to the training set, while the remainder are designated for the testing set. This mechanism facilitates a more straightforward learning process for models to grasp the human rating standards. Conversely, a high degree of variance in human-assigned scores suggests that the paper (or idea) is controversial, rendering the learning of standards from such data as potentially futile.

\subsection{Layer Selection}
As claimed by \citet{geva2021transformer}, lower layers of LLMs tend to capture shallow data patterns, while upper layers contain more semantic knowledge. This hierarchical processing of information within LLMs suggests that the utility of representations may vary across layers. Further to this, RepE~\cite{zou2023representation} explores the relationship between layer depth and performance in utility estimation tasks, finding that middle-layer representations often yield the highest accuracy.

Inspired by these findings, our approach involves identifying the optimal layer within an LLM that provides the most effective representations for constructing an accurate evaluator. We hypothesize that a specific intermediate layer may offer the ideal balance between capturing fundamental linguistic features and the nuanced semantic understanding necessary for assessing the quality of scientific ideas. Our experiments are thus tailored to pinpoint this layer by analyzing the performance of given data across all layers. Then, we leverage its representations to enhance the performance of our idea evaluation framework.

\subsection{Token Selection}\label{sec_token_selection}
Considering that a manuscript $d_i$ is composed of $l$ sequential tokens, the semantic information of these token representations varies significantly. Due to the fact that most LLMs are auto-regressive models, the last token aggregates the attention information of all previous tokens~\cite{zou2023representation}. With a slight abuse of notation, by default, we use the last token representation $Rep(d_i, -1) \in \mathbb{R}^{m}$ to symbolize the entirety of the manuscript $d_i$.

Nevertheless, when dealing with lengthy input texts, such as full-text manuscript $d_i$, there are two issues with the default approach. For one thing, memory optimization mechanism such as vLLM~\cite{kwon2023efficient} should be adopted to prevent GPU from running out of memory. For another thing, the representation of the last token may become diluted or overly abstracted owing to the extensive accumulation of attention, potentially leading to a loss of specific semantic details pertinent to the overall idea assessment. To address these issues, we explore alternative strategies for token selection that aim to maintain the richness of semantic information while ensuring computational feasibility.

We consider a manuscript $d_i$ to be a composition of distinct sections. We select the last token representations from each section, and concatenate these to form a composite representation. The approach allows us to capture the essence of each section. Formally, if a manuscript $d_i$ is divided into $r$ sections, and $Rep(d_{i,j}, -1)$ represents the last token of the $j^{th}$ section, then the combined representation $Rep(d_i)$ is given by:

\begin{equation}
Rep(d_i) = \bigoplus_{j=1}^{r} Rep(d_{i,j}, -1),
\end{equation}
where $\bigoplus$ denotes the concatenation operation, and $Rep(d_i)$ is in $\mathbb{R}^{r \times m}$. Similarly, we can take into account the full length of the manuscript and divide it into equidistant segments based on a predefined length to obtain $Rep(d_i)$. By experimenting with these strategies, we aim to refine our approach to token selection and optimize the representation of manuscript for more accurate idea assessment.

\subsection{Evaluator Training}
In this part, we use the pre-processed $Rep(d_i)$ to train an idea evaluator $A_c$. Let $s_i$ be the average score given by humans for manuscript $d_i$, reflecting its overall quality according to the criterion $c$. The average score serves as the ground truth in our training process. The evaluator $A_c$ is instantiated as an Multilayer Perceptron (MLP) with one hidden layer. The MLP is tasked with learning the mapping from the representation space to the scalar scores, which takes as input the representation $Rep(d_i)$ for each manuscript $d_i$ and outputs a predicted score $\hat{s}_i$. To optimize all parameters of the MLP, we employ the Mean Squared Error (MSE) loss function:

\begin{equation}
    \mathcal{L} = \frac{1}{n} \sum_{i=1}^{n} (\hat{s}_i - s_i)^2.
\end{equation}

By minimizing $\mathcal{L}$, the MLP learns to approximate the human-assigned scores as closely as possible. Through this training process, we aim to calibrate the evaluator $A_c$ such that it can reliably predict the value of new or unseen ideas based on their textual representations.

\begin{table*}[htb]
\centering
\small
\tabcolsep 0.146in
\begin{tabular}{lllllllll}
\toprule[1pt]
\multicolumn{1}{l|}{\multirow{3}{*}{\textbf{Method}}} & \multicolumn{4}{c|}{\textbf{ICLR23-low-std}}                                                                                                     & \multicolumn{4}{c}{\textbf{ICLR23-all}}                                                                                                         \\ \cmidrule{2-9} 
\multicolumn{1}{l|}{}                                 & \multicolumn{2}{c}{\textbf{train: 5\%}}                                   & \multicolumn{2}{c|}{\textbf{train: 30\%}}                                  & \multicolumn{2}{c}{\textbf{train: 5\%}}                                   & \multicolumn{2}{c}{\textbf{train: 30\%}}                                  \\ \cmidrule{2-9}
\multicolumn{1}{l|}{}                                 & \multicolumn{1}{c}{\textbf{corr}} & \multicolumn{1}{c}{\textbf{layer}} & \multicolumn{1}{c}{\textbf{corr}} & \multicolumn{1}{c|}{\textbf{layer}} & \multicolumn{1}{c}{\textbf{corr}} & \multicolumn{1}{c}{\textbf{layer}} & \multicolumn{1}{c}{\textbf{corr}} & \multicolumn{1}{c}{\textbf{layer}} \\ \midrule
\multicolumn{9}{c}{\textbf{LLM Generation}}                                                                                                                                                                                                                                                                                                                \\ \midrule
\multicolumn{1}{l|}{LLaMA-2-Full-SFT}                 & N/A                               & N/A                                & N/A                               & \multicolumn{1}{l|}{N/A}            & N/A                               & N/A                                & N/A                               & N/A                                \\
\multicolumn{1}{l|}{LLaMA-2-LoRA-SFT}                 & -0.0513                           & N/A                                & 0.0820                            & \multicolumn{1}{l|}{N/A}            & 0.0634                            & N/A                                & 0.0692                            & N/A                                \\
\multicolumn{1}{l|}{Baichuan-2-LoRA-SFT}              & 0.1391                            & N/A                                & 0.2054                            & \multicolumn{1}{l|}{N/A}            & 0.1413                            & N/A                                & 0.1867                            & N/A                                \\
\multicolumn{1}{l|}{GPT-3.5-turbo}                    & 0.1290                            & N/A                                & 0.1375                            & \multicolumn{1}{l|}{N/A}            & 0.0874                            & N/A                                & 0.0719                            & N/A                                \\ \midrule
\multicolumn{9}{c}{\textbf{LLM Representation}}                                                                                                                                                                                                                                                                                                            \\ \midrule
\multicolumn{1}{l|}{BERT}                             & 0.1986                            & -3                                 & 0.2515                            & \multicolumn{1}{l|}{-4}             & 0.1907                            & -1                                 & 0.2326                            & -1                                 \\
\multicolumn{1}{l|}{SciBERT}                          & 0.2677                            & -3                                 & 0.3314                            & \multicolumn{1}{l|}{-3}             & 0.2447                            & -2                                 & 0.2584                            & -3                                 \\
\multicolumn{1}{l|}{RePE-with-prompt}                 & 0.0820                            & -31                                & 0.0993                            & \multicolumn{1}{l|}{-31}            & 0.0738                            & -21                                & 0.0880                            & -31                                \\
\multicolumn{1}{l|}{RePE-no-prompt}                   & 0.1020                            & -1                                 & 0.0605                            & \multicolumn{1}{l|}{-31}            & 0.0635                            & -2                                 & 0.0508                            & -2                                 \\
\multicolumn{1}{l|}{\textbf{Ours}}                    & \textbf{0.3441}                   & -20                       & \textbf{0.3880}                            & \multicolumn{1}{l|}{-9}             & \textbf{0.2783}                            & -1                                 & \textbf{0.3366}                            & -4                                 \\ \midrule
\multicolumn{9}{c}{\textbf{Human Evaluation}}                                                                                                                                                                                                                                                                                                              \\ \midrule
\multicolumn{1}{l|}{Human}                    & 0.8175                            & N/A                                & 0.7648                            & \multicolumn{1}{l|}{N/A}            & 0.4174                            & N/A                                & 0.3290                            & N/A                                \\ \bottomrule[1pt]
\end{tabular}
\caption{Spearman correlations with humans of different methods on ICLR23 datasets. N/A in \textit{corr} column means its corresponding $pvalue > 0.05$. There is no need for LLM Generation baselines to select layers. The human performance is evaluated by randomly selecting one score from the human-rated list against other reviews.}
\label{table_cmp}
\end{table*}

\section{Experiments}
This section presents a series of experiments to verify the performance of LLMs in the task of quantitative evaluation of ideas. Our main focus is on the mean value of the criteria \textbf{overall quality}, which is used as the training objective for the idea evaluator. Through our released benchmark and the experimental methodology, we answer the following four research questions (RQs):

\begin{itemize}
    \item \textbf{RQ1}: To what extent do the representations from LLMs correlate with human judgements in the evaluation of scientific ideas? Additionally, is the LLM generation method suitable for this task?
    \item \textbf{RQ2}: What is the impact of choosing different layers and tokens for LLM representations on the performance of idea evaluation?
    \item \textbf{RQ3}: How significantly does the consistency of human judgements influence the performance of LLM representations in this context?
    \item \textbf{RQ4}: How does the size of training set impact the correlation between $A_c$ evaluations and human judgments in idea assessment?
\end{itemize}

\subsection{Baselines}
We categorize the baselines into three distinct groups: LLM Generation, LLM Representation, and Human Evaluation. The first category involves LLMs generating numerical scores in response to textual descriptions of ideas. This category includes models such as GPT-3.5-turbo, LLaMa-2-7b-base~\cite{touvron2023llama}, and Baichuan-2-7b-base~\cite{yang2023baichuan}, which are fine-tuned using techniques like LoRA~\cite{hu2021lora} or with full parameter updates. The prompts we choose are presented in Appendix~\ref{sec_prompts}. In the LLM Representation category, we evaluate models like BERT~\cite{kenton2019bert}, SciBERT~\cite{beltagy2019scibert}, and RePE~\cite{zou2023representation}. For BERT and SciBERT, we also apply our proposed framework to quantify the value of ideas, with the primary distinction being in the token selection strategy. Specifically, we used the $[CLS]$ token as the representation of an idea, and if the length of a section exceeds 512 tokens, we will divide it into equidistant subsections to apply the token selection strategy for BERT-like models. Moreover, we also analyze the performance of human evaluators through randomly selecting one score from the human-rated list against other scores.

\subsection{Training Settings and Evaluation Details}\label{sec_setting}
In our implementation, our method employs LLaMA-2-7b-base as the foundational model. In order to make our experiments more solid and validate our framework is model-agnostic, we also use Baichuan-2-7b-base as the base model, the results of which are provided in Appendix~\ref{sec_expaned_results}. We use the grid search to find appropriate sets of hyper-parameters for baselines and our proposed method. For the configuration of the MLP evaluator, we choose a batch size of 32, a hidden layer dimension of 1024, a learning rate of 0.001, a dropout rate of 0.2, and employ the Adam optimizer. We limit the training to 20 epochs. More detailed settings are documented in Appendix~\ref{sec_parameters}. Each experiment is executed three times with random initializations, and the mean results are reported. We use the results of the training set for model selection. All experiments are first conducted to evaluate the efficacy of our framework using the \textbf{abstracts of papers} for all research questions. We also explore the effects of using the \textbf{full texts of papers} as the training inputs for the token selection in RQ2.

To gauge the alignment of scores generated by various methods with human-assigned scores, we report a widely-used metric called \textit{Spearman Correlation}~\cite{spearman1961proof}. The correlation $corr$ with human is defined as:
\begin{equation}
    corr=\rho([s_1, s_2, ..., s_n], [\hat{s}_1, \hat{s}_2, ..., \hat{s}_n]),
\end{equation}
where $\rho$ is the Spearman Correlation function, $s_i$ is the average score given by humans, and $\hat{s}_i$ is the $A_c$ predicted score for $d_i$ in the testing set. Since Spearman correlation is invariant under affine transformations, we also provide score distribution and the absolute error between human-rated scores and our LLM representation scores in Section~\ref{score_dist} and Appendix~\ref{domain_analysis}.

\subsection{Comparative Experiments (RQ1)}
According to the principle of consistency sorting in Section~\ref{c_sorting}, we construct training sets using the top 5\% and top 30\% ratios from ICLR23-low-std and ICLR23-all datasets respectively to preliminarily exclude the influence of dataset proportion on the conclusion, and take the rest of each dataset as the testing set. Table~\ref{table_cmp} shows the Spearman correlations with humans of different methods on these two datasets. We also provide indexes for the layers with the highest correlation. 

It can be observed that our proposed method achieves the best performance in all settings, where the performance on ICLR23-all is \textbf{at most 30\%} higher than the second best method SciBERT. As expected, the correlation of ICLR23-low-std among human scores is close to 1, which is attributed to our data partitioning strategy. It should be noted that the correlation of our method on ICLR23-all dataset exceed the result of humans, when the training ratio is 30\%, proving the feasibility of our method and its potential ability to be applied to real-world review scenarios. Moreover, in terms of different layers' performance, the middle and back layers of most models may achieve better results.

For the LLM Generation baselines, the fine-tuned LLaMA-2 is worse than Baichuan-2, especially for the LLaMA-2-Full-SFT, fine-tuned with full parameters, lacking the capability of effective evaluation since its pvalue > 0.05. Due to the inability of GPT-3.5 being fine-tuned, we adopt the zero-shot setting, which is only for sketchy reference. We also try k-shot setting for GPT-3.5, but it only generates the most frequent scores from the given examples. Overall, the LLM Generation methods are not competent for the quantitative evaluation of ideas. By analyzing the generated results, we believe there are two possible reasons. One is that the amount of training data is relatively small, and models are prone to overfitting. Apart from that, LLM generation is not sensitive to digital numbers, and the semantic knowledge is hidden in its representations, which should be guided through appropriate means.

Furthermore, our experiment studies the degree of consistency between the predicted score and that assigned by the human reviewer whose score most closely match the predicted one. As depicted in Table~\ref{table_vs_human}, the ICLR23-all dataset exhibits a higher consistency with the closest human-rated scores compared to the ICLR23-low-std dataset. This suggests that, despite the higher variance in human scores of the ICLR23-all dataset, our proposed method is adept at mirroring the evaluation of the most similar human reviewer. 

\begin{table}[!htb]
\centering
\small
\tabcolsep 0.13in
\renewcommand{\arraystretch}{1.2}
\begin{tabular}{llcc}
\toprule[1pt]
\multirow{2}{*}{\textbf{Dataset}} & \multirow{2}{*}{\textbf{Method}} & \multicolumn{2}{l}{\textbf{ Training Ratio}} \\
                                  &                                  & \textbf{5\%}         & \textbf{30\%}        \\ \hline
\multirow{2}{*}{ICLR23-low-std}   & SciBERT                          & 0.4299               & 0.5093               \\
                                  & \textbf{Ours}                    & \textbf{0.5469}      & \textbf{0.5605}      \\ \hline
\multirow{2}{*}{ICLR23-all}       & SciBERT                          & 0.6381               & 0.5112               \\
                                  & \textbf{Ours}                    & \textbf{0.6462}      & \textbf{0.5617}      \\ \bottomrule[1pt]
\end{tabular}
\caption{Spearman correlations with the closest human-rated score.}
\label{table_vs_human}
\end{table}

\subsection{Score Distribution (RQ1)}\label{score_dist}

\begin{figure}[ht]
  \centering
  \begin{minipage}[b]{0.21\textwidth}
    \includegraphics[width=\textwidth]{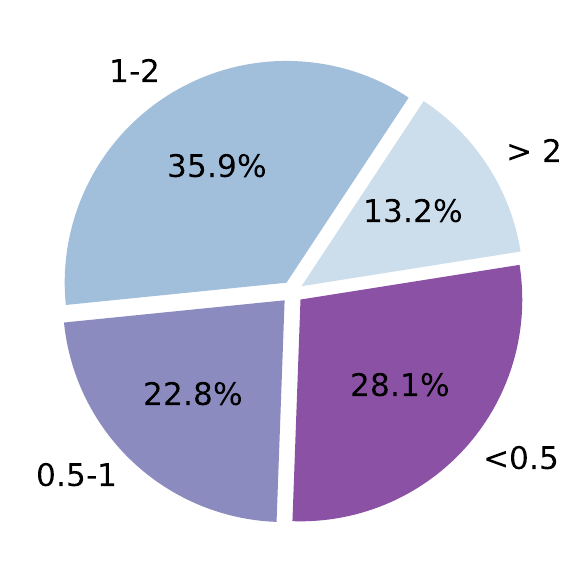}
  \end{minipage}
  \hfill
  \begin{minipage}[b]{0.26\textwidth}
    \includegraphics[width=\textwidth]{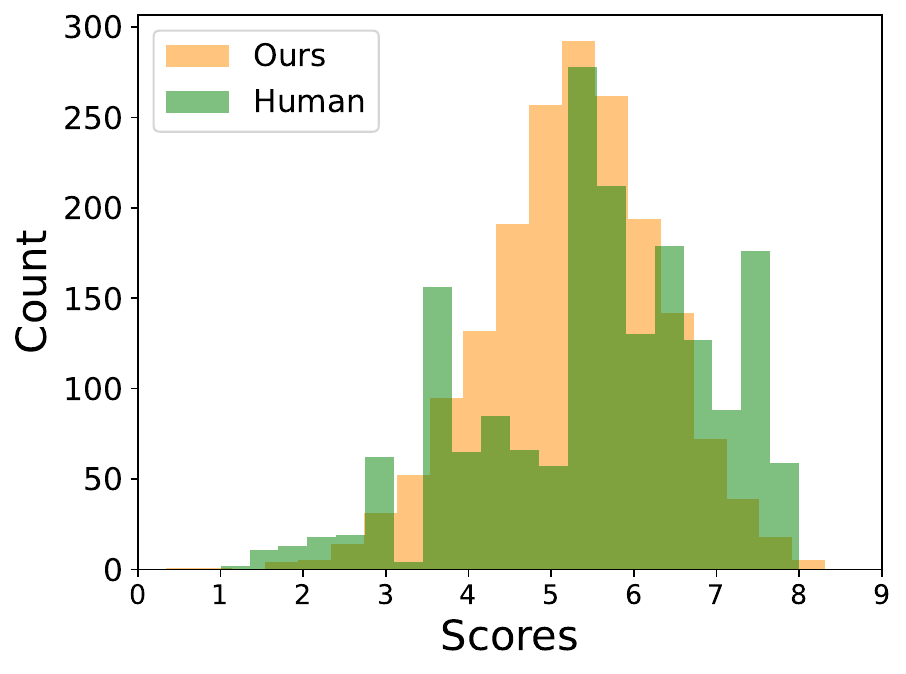}
  \end{minipage}
  \caption{The difference (absolute error) between human-rated scores and the predicted scores (left subfigure). The distributions of human-rated scores and the predicted scores (right subfigure).}
  \label{fig_score_distribution}
\end{figure}

We also examine the difference (absolute error) between human-rated scores and the predicted scores on ICLR23-low-std dataset. The results are shown in the pie chart of Figure~\ref{fig_score_distribution}. We can see that \textbf{86.8\%} paper scores generated by our method are close to the human-rated scores, where the differences between them are lower than 2. Additionally, the distributions are shown in the right part of Figure~\ref{fig_score_distribution}. The distribution of scores predicted by our idea evaluator is normal distribution as expected while the human reviewers tend to give more higher or lower scores. More analysis can be found in Appendix~\ref{domain_analysis}.

\subsection{Influence of Layer Selection (RQ2)}

\begin{figure}[htb]
    \centering
    \includegraphics[width=1.0\linewidth]{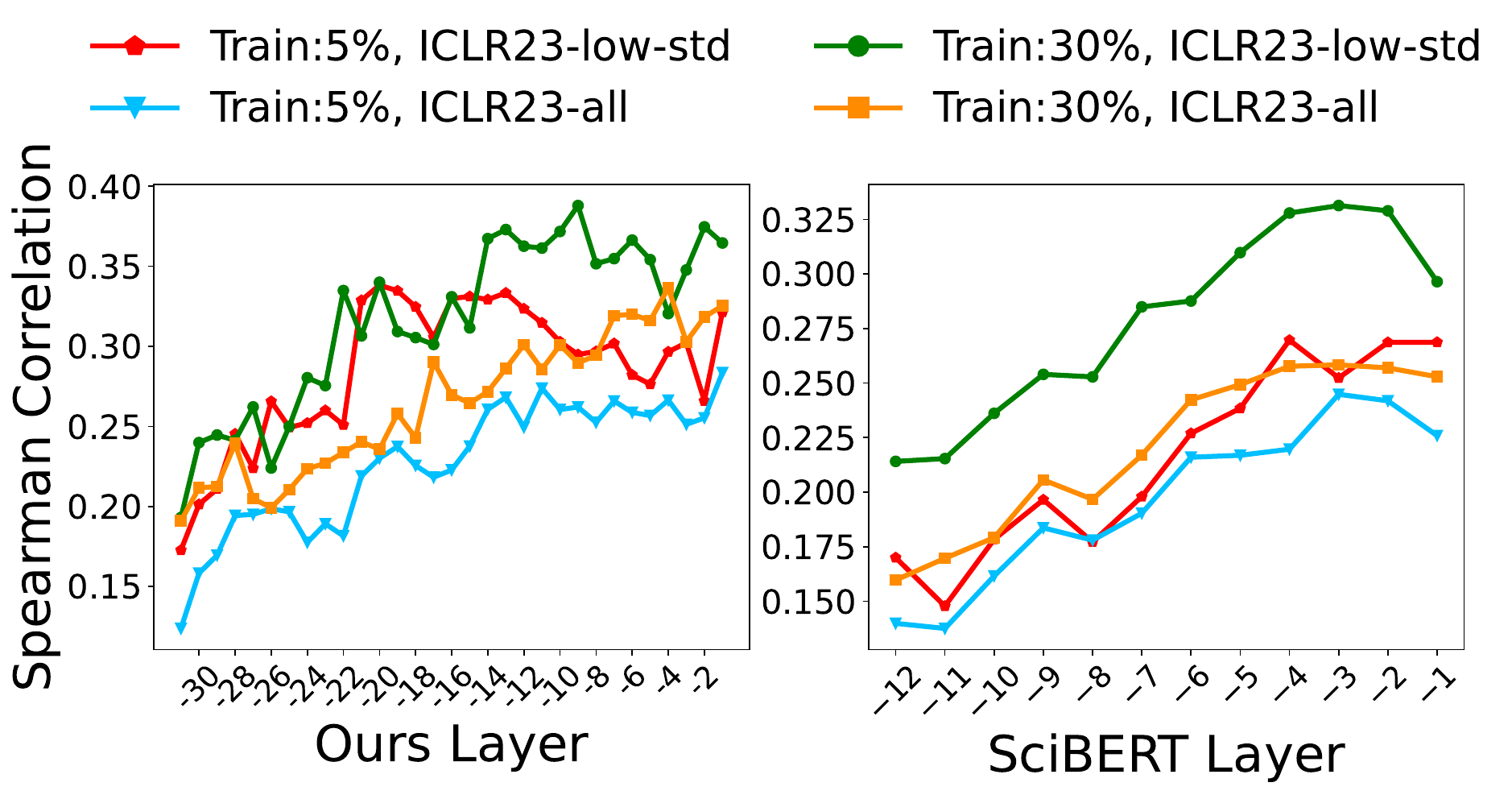}
    \caption{Spearman correlations of different layers in LLM and SciBERT.}
    \label{fig_layer}
\end{figure}

We analyzed the representational efficacy across various layers of LLM and SciBERT. As illustrated in Figure~\ref{fig_layer}, it is evident that for both LLM and SciBERT, the representations from the middle to later layers outperform those from other layers. Obviously, the very last layers do not typically yield the best performance. This may be attributed to the specific semantic information encapsulated in different layers. The last layer is inherently born to facilitate generation tasks, rather than tasks that require more discriminative capabilities, like classification or regression. 

Given the nuanced role of layer-specific representations in the context of assessing the merit of scientific ideas, we propose a layer selection of representations for the task at hand. Specifically, we advocate for the utilization of representations from the layers situated in the last one-third of the model's depth. Such choice is informed by the empirical evidence suggesting that these layers strike a balance between retaining rich semantic content and providing the necessary abstraction for discriminative tasks. 

\begin{figure*}[htb]
    \centering
    \includegraphics[width=1.0\linewidth]{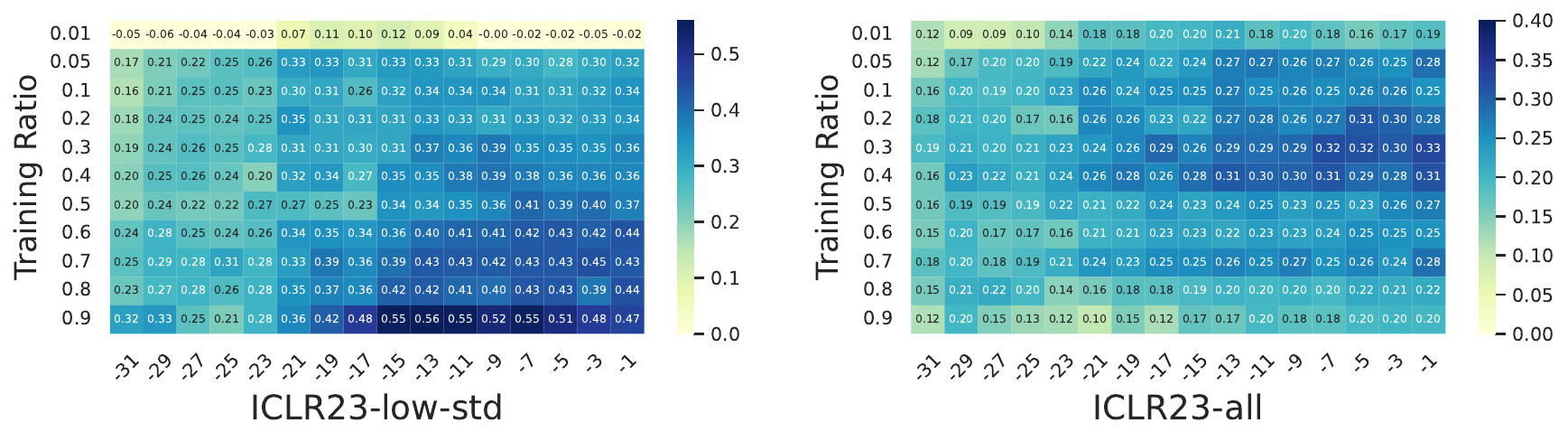}
    \caption{Spearman correlations varying with layers under different training ratios on ICLR23 datasets.}
    \label{fig_training_ratio}
\end{figure*}

\subsection{Influence of Token Selection (RQ2)}
In this part, we use the ICLR23-all dataset to investigate the influence of token selection. In terms of the paper abstract inputs, we first test the correlation results of using the last token. Subsequently, we expand our scope to include both the middle and last tokens (middle + last token) of the input text. The findings, as presented in Table~\ref{table_token_selection}, indicate that solely relying on the last token yields superior results compared to combining it with the middle token. The latter approach appears to introduce a surplus of redundant information that may hinder the downstream performance of evaluator.

\begin{table}[!htb]
\centering
\small
\tabcolsep 0.17in
\renewcommand{\arraystretch}{1.2}
\begin{tabular}{lcc}
\toprule[1pt]
\multirow{2}{*}{\textbf{Token Selection Strategy}} & \multicolumn{2}{l}{\textbf{ Training Ratio}} \\
                                                   & \textbf{5\%}       & \textbf{30\%}       \\ \hline
\multicolumn{3}{l}{\textbf{Abstract}}                                                         \\
last token                                         & 0.2783              & 0.3366             \\
middle + last token                                & 0.2162              & 0.2772             \\ \hline
\multicolumn{3}{l}{\textbf{Full-text}}                                                        \\
segment tokens                                    & 0.1306              & 0.2597             \\
section tokens                                     & 0.3258              & 0.3821             \\ 
\bottomrule[1pt]
\end{tabular}
\caption{Spearman correlations of different token selection strategies.}
\label{table_token_selection}
\end{table}

When dealing with the full text of papers, we implement two token selection strategies outlined in Section~\ref{sec_token_selection}: the amalgamation of last tokens from equidistant segments and the aggregation of last tokens from distinct paper sections. The experimental outcomes suggest that merging last tokens from all segments does not effectively capture the necessary semantic content. This is largely due to the fact that many segments are incomplete sentences, arbitrarily truncated from the original texts. On the other hand, the section-based strategy successfully compiles comprehensive information from each complete section. Overall, we recommend using the representations of the last tokens or section-based strategy to train the idea evaluator. 


\subsection{Analysis of Training Set (RQ3 \& RQ4)}
Figure~\ref{fig_training_ratio} shows the Spearman correlations varying with layers under different training ratios on ICLR23 datasets. Due to space limitations, we only present the results of odd numbered layers. See Appendix~\ref{sec_expaned_results} for more results. The manuscript papers are sorted in ascending order of variance in their corresponding human-rated scores to partition datasets according to the consistency sorting in Section~\ref{c_sorting}.

For ICLR23-low-std dataset, the human-rated scores in the dataset are highly consistent, and it is observed that the Spearman correlation tends to improve in tandem with increases in the training set size. Notably, when the proportion of data used for training surpasses the 50\% threshold, the correlation between the scores predicted by our idea evaluator and those assigned by human experts becomes moderate, exceeding 0.4. Furthermore, our analysis reveals that even a relatively small subset of the data (with a training ratio of 5\%) is capable of yielding positive performance. 


As to ICLR23-all dataset, the outcomes indicate that an increase in the volume of training data does not necessarily correspond to a higher alignment with human evaluations in the testing set. The phenomenon can be attributed to the diminishing consistency of human scores as the dataset expands; that is, the variance in human-assigned scores grows with the size of the dataset. It becomes evident that while a larger training set generally provides more information, it also introduces a greater diversity of human judgment, which may not always be conducive to improving the ability of evaluator to mimic human scoring behavior.


\section{Conclusion and Future Work}
The study focuses on the quantitative evaluation of scientific ideas. We have reviewed existing methodologies for paper and idea evaluation and have broken new ground by focusing on the quantitative aspect of idea evaluation. Specifically, we first introduce a comprehensive benchmark dataset, accessible to the research community. Then, we develop a new framework that leverages the token representations of specific layers in LLM to quantify the value of ideas. Through rigorous experiments, we demonstrate that LLM representations correlate more strongly with human judgments compared to generative text outputs. Additionally, in our benchmark, the predicted scores of more than 80\% papers are close to human-rated scores. In the future, we will broaden the scope of our research to encompass diverse disciplines with balanced data ratios, including the exact and social sciences, to further validate and refine our evaluative framework.

\clearpage

\section*{Limitations}

\subsection*{Discipline} The scope of our research is confined to the field of computer science, which may restrict the broader applicability of our framework. The generalization performance of our model across different scientific disciplines remains an open question. Future research endeavors should aim to adapt and validate the framework in diverse fields, ranging from the exact sciences to the humanities.

\subsection*{Criterion} Our experiments have primarily focused on the overall quality score of manuscript papers, which is a composite yet somewhat abstract. Important aspects such as the correctness of the presented work and its novelty are equally critical in determining a paper's impact and significance. In forthcoming studies, we plan to dissect these individual criteria, developing a more granular approach to idea evaluation.

\subsection*{Model Scale} The impact of model scale on performance is an aspect that has not been extensively explored in our research. The performance of LLMs is often closely tied to the number of parameters they contain; thus, models with different sizes may yield different results in the task of idea evaluation. Larger models may have the capacity to encode more nuanced representations of text, potentially leading to more accurate assessments of scientific ideas. Conversely, they may also introduce complexities that do not necessarily translate to better performance, such as overfitting or increased computational costs. The trade-offs between model size, accuracy, and efficiency are still an area ripe for exploration.

\section*{Ethics Statement}
The dataset used in our study consists of publicly available academic papers. We have ensured that all data was collected and handled in a manner that respects the privacy and intellectual property rights of the authors. No personal data was used, and all information is attributed to its original source.

We are committed to transparency in our research process. To this end, we have made our benchmark dataset publicly available and have provided detailed descriptions of our methodologies and experimental setups to facilitate reproducibility by other researchers.

We recognize the importance of the human element in the evaluation of scientific ideas. Our framework is designed to assist, rather than replace, human judgment. We believe that the most effective use of our model is as a tool to support and enhance the work of human reviewers, not to supplant them.

\bibliography{custom}

\clearpage

\appendix

\section{Prompts for LLM Generation}\label{sec_prompts}
For the baselines involving LLM generation, we design a prompt to elicit the evaluation of a manuscript paper's \textbf{overall quality} based on its abstract. The prompt, which demonstrated optimal performance when applied to GPT-3.5-turbo, is structured as follows:

\begin{myshaded}
\noindent \textbf{Evaluate} the quality (overall quality score) of the following manuscript paper based on its abstract.\\ \\
\textbf{title}: \{title\}\\ \\
\textbf{abstract}: \{abstract\}\\ \\
The score should be between 1 and 10, with 1 being the lowest and 10 being the highest. Just output your score, no more other words.
\end{myshaded}

It is important to note that while this prompt is most effective for GPT-3.5-turbo, its influence on the performance of other fine-tuned models, such as LLaMA-2 and Baichuan-2, is less pronounced. These models have been trained to adapt to the distribution of scores in the training set, which mitigates the impact of the prompt's phrasing on their generative capabilities.

\section{Hyper Parameters}\label{sec_parameters}
We first declare that the reason for using \textbf{LLaMA-2 instead of LLaMA-3} or other updated models is because we are concerned that new models may be pretrained using papers from ICLR23, resulting in a data leakage problem.

\begin{table}[!htb]
\centering
\tabcolsep 0.44in
\begin{tabular}{ll}
\toprule[1pt]
\textbf{Parameter} & \textbf{Value} \\ \hline
learning rate      & 2e-5           \\
epoch              & 3              \\
weight decay       & 0              \\
warmup ratio       & 0.03           \\
bf16               & True           \\ \bottomrule[1pt]
\end{tabular}
\caption{Hyper-parameters of LLaMA-2-Full-SFT.}
\label{table_hyper_full}
\end{table}

We detail the hyper-parameters for the LLM generation baseline, LLaMA-2-Full-SFT, in Table~\ref{table_hyper_full}. Additionally, the hyper-parameters for models trained with the LoRA technique, specifically LLaMA-2-LoRA-SFT and Baichuan-2-LoRA-SFT, are outlined in Table~\ref{table_hyper_lora}.

\begin{table}[!htb]
\centering
\tabcolsep 0.08in
\begin{tabular}{lll}
\toprule[1pt]
\textbf{Parameter} & \textbf{LLaMA-2} & \textbf{Baichuan-2} \\ \hline
learning rate      & 2e-5             & 2e-5                \\
epoch              & 10               & 10                  \\
weight decay       & 0                & 0                   \\
warmup ratio       & 0                & 0                   \\
bf16               & True             & True                \\
LoRA modules       & q\_proj, v\_proj & W\_pack             \\
LoRA r             & 8                & 16                  \\
LoRA alpha         & 16               & 32                  \\
LoRA dropout       & 0.05             & 0.1                 \\ \bottomrule[1pt]
\end{tabular}
\caption{Hyper-parameters of LLaMA-2-LoRA-SFT and Baichuan-2-LoRA-SFT.}
\label{table_hyper_lora}
\end{table}

For the representation-based evaluation method RePE~\cite{zou2023representation}, we employ Principal Component Analysis (PCA) as the embedding evaluation mechanism, in line with the recommendations provided in the original paper.

\section{Expanded Results}\label{sec_expaned_results}

\begin{figure*}[htb]
    \centering
    \includegraphics[width=1.0\linewidth]{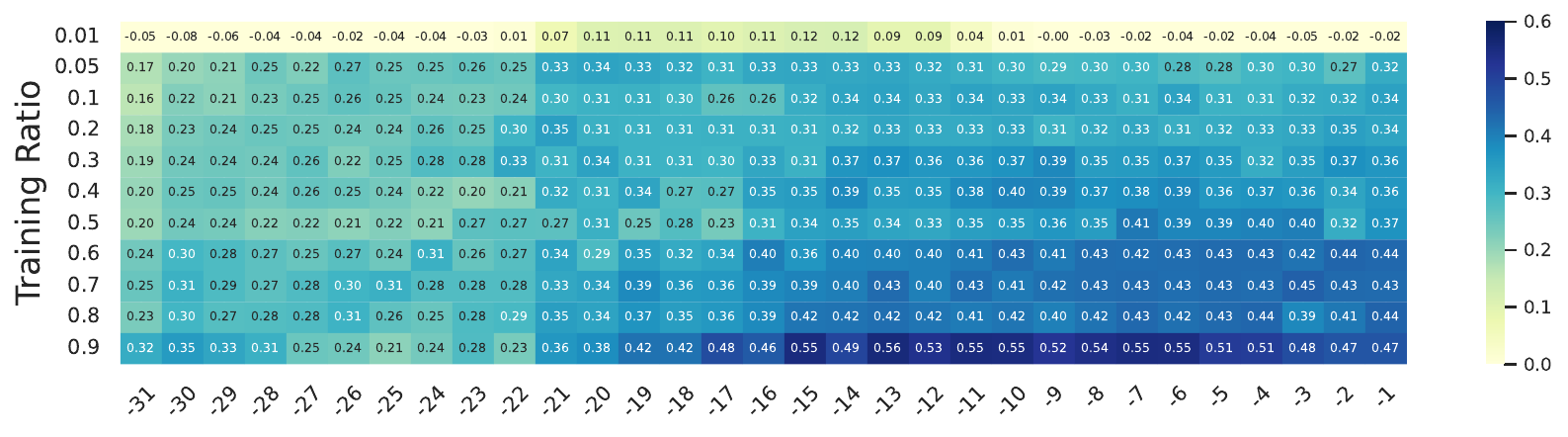}
    \caption{Spearman correlations varying with layers under different training ratios of \textbf{LLaMA-2-7b-base} on \textbf{ICLR23-low-std} dataset.}
    \label{fig_result_llama_low_std}
\end{figure*}

\begin{figure*}[htb]
    \centering
    \includegraphics[width=1.0\linewidth]{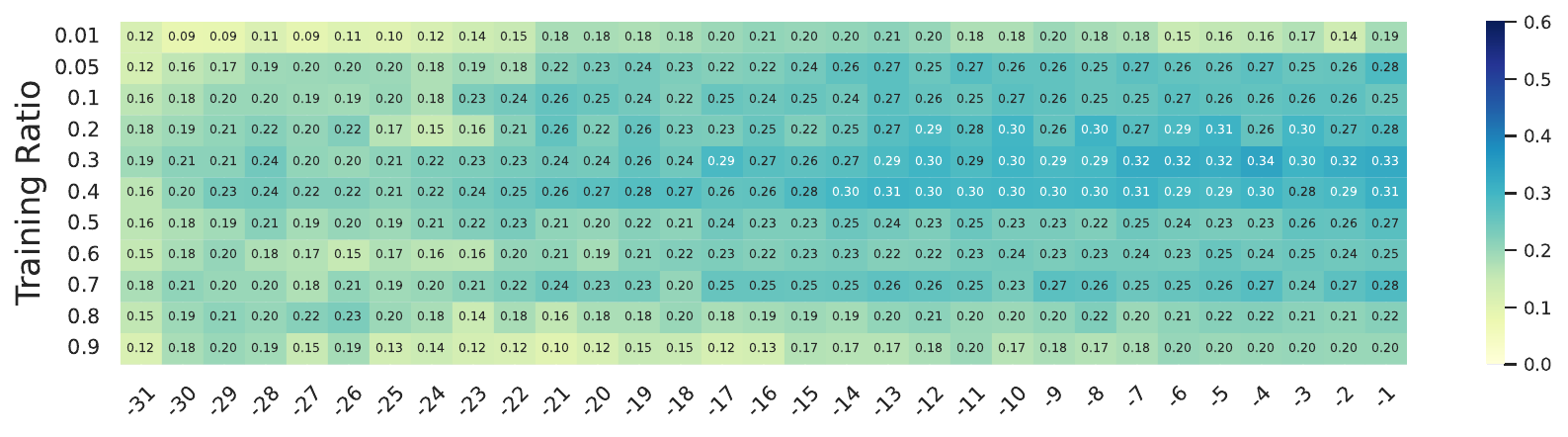}
    \caption{Spearman correlations varying with layers under different training ratios of \textbf{LLaMA-2-7b-base} on \textbf{ICLR23-all} dataset.}
    \label{fig_result_llama_all}
\end{figure*}

Building upon the experimental setup detailed in Section~\ref{sec_setting}, we examine the performance across \textbf{all layers} of the foundational model, LLaMA-2-7b-base. The detailed Spearman correlation results, which consider the full spectrum of layers under various training ratios, are illustrated in Figure~\ref{fig_result_llama_low_std} for the ICLR23-low-std dataset and in Figure~\ref{fig_result_llama_all} for the ICLR23-all dataset.

\begin{figure*}[htb]
    \centering
    \includegraphics[width=1.0\linewidth]{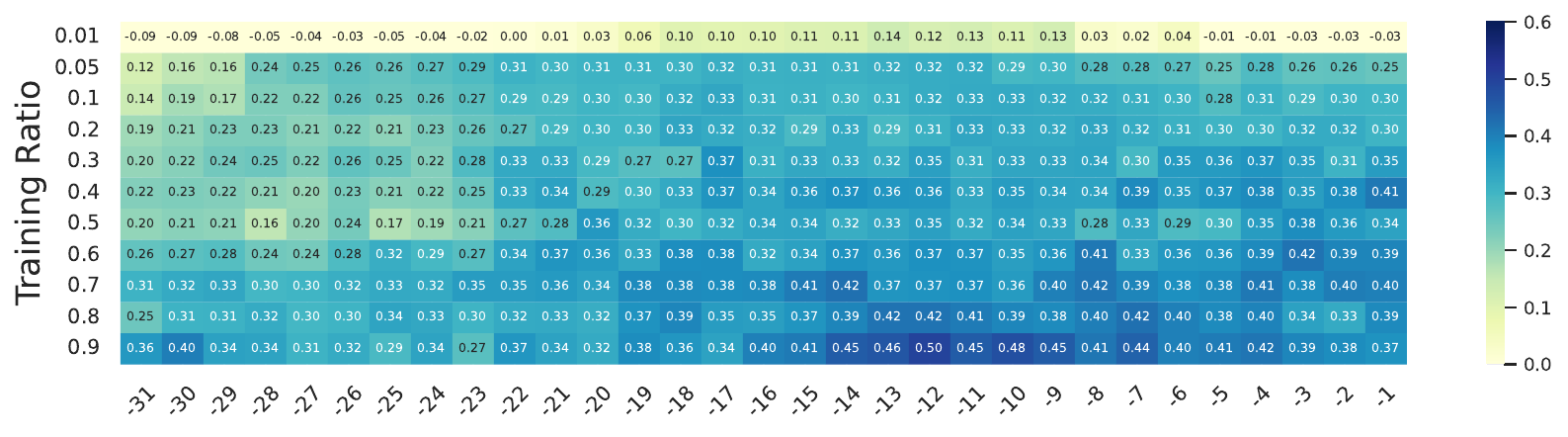}
    \caption{Spearman correlations varying with layers under different training ratios of \textbf{Baichuan-2-7b-base} on \textbf{ICLR23-low-std} dataset.}
    \label{fig_result_baichuan_low_std}
\end{figure*}

\begin{figure*}[htb]
    \centering
    \includegraphics[width=1.0\linewidth]{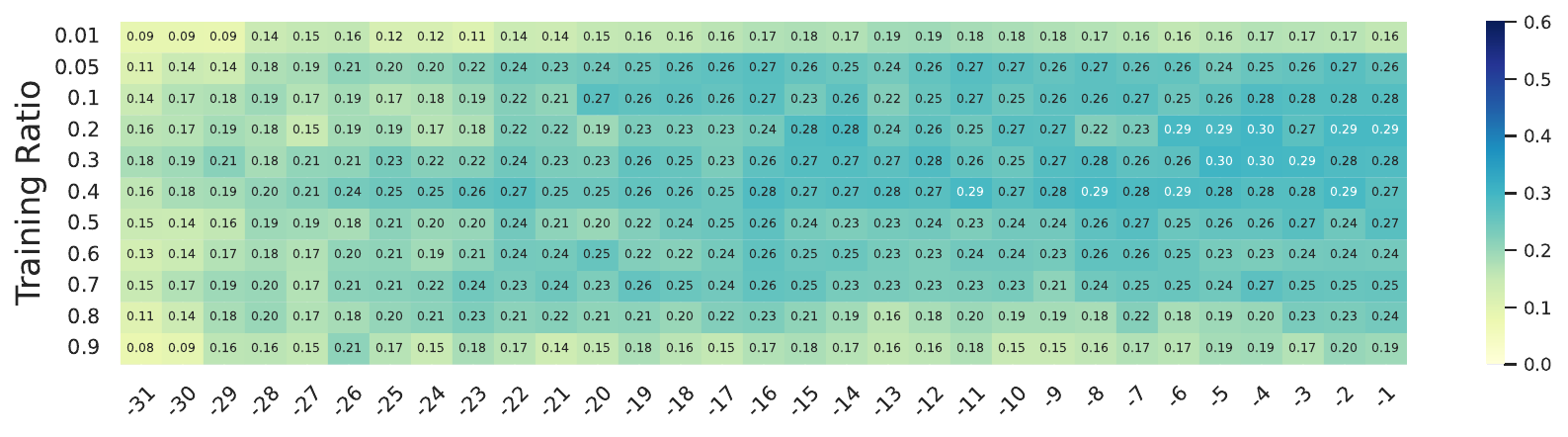}
    \caption{Spearman correlations varying with layers under different training ratios of \textbf{Baichuan-2-7b-base} on \textbf{ICLR23-all} dataset.}
    \label{fig_result_baichuan_all}
\end{figure*}

In our pursuit to validate the robustness of our findings, we conducted parallel experiments using Baichuan-2-7b-base as an alternative base model. The corresponding Spearman correlation results are depicted in Figure~\ref{fig_result_baichuan_low_std} and Figure~\ref{fig_result_baichuan_all}. The patterns observed with Baichuan-2-7b-base are found to be in harmony with those from the LLaMA-2-7b-base model, lending credence to the consistency and reliability of our conclusions.

The experiments across different models not only reinforces the validity of our initial observations but also suggests that the underlying phenomena we have identified are model-agnostic to a certain extent. Such findings are indicative of the potential generalizability of our framework, hinting at its applicability across a variety of LLMs. Future work may delve deeper into the comparative analysis of additional models, further expanding our understanding of the relationship between the base model and the efficacy of idea evaluation.

\section{Case Study}\label{sec_case_study}
We list four cases to show the performance and drawbacks of our method in Table~\ref{table_case_study}. All these cases are selected from the domain of reinforcement learning. The first two cases are correctly predicted, and the human-rated scores and LLM representation scores are very close. As to the third case, although our method gives an overestimated score, the final score is not enough to make it acceptable. The fourth case is underestimated. One possible reason is that in such cases, our method may lack more contextual information to make a decision, such as tables and figures in papers, which is also something we need to consider in the future.

\section{Domain Analysis}\label{domain_analysis}
To see how our method rates ideas on popular topics or less trendy domains, we analyze the score distributions and differences between human-rated scores and LLM representation scores in 14 domains divided by ICLR-2023 program committee. The results are shown in Table~\ref{table:score_details}. On the whole, the differences in mean scores of most domains are less than 10\%. However, there are three domains (\textit{Theory, Neuroscience and Cognitive Science, Infrastructure}) where the mean values of human-rated scores are relatively higher than average, and the differences exceed 10\%. We believe these three domains are distinguished from other domains since others frequently focus on \textit{Learning and Optimization}, which makes our evaluator overfitting on these data. Therefore, it is necessary to train a domain-specific evaluator, while also maintaining a balance in the content of the dataset. We will address these issues in our future work.

\section{Frequently Asked Questions}

\subsection*{Numerical Processing Limitations in LLMs}
A critical issue faced by LLMs is their inherent difficulty in processing numerical data, such as digits. This limitation stems from the finite-sized vocabulary and tokenization strategies used by these models, affecting both encoder and decoder architectures. This impacts the ability to perform tasks that require precise numerical understanding. Therefore, we leverage the deep, contextual representations within LLMs to quantify the value of scientific ideas. These representations encapsulate rich semantic and contextual information that extends beyond the superficial token sequences. 

Our approach diverges significantly from the strategy of merely adding task-specific heads to the model. Instead, our approach involves a strategic selection of layers and tokens. By leveraging the hierarchical processing capabilities of LLMs, we can harness the most relevant and informative features for idea evaluation. This approach contrasts with using the entire weight set of the LLM (adding head to LLM), which might not be as efficient or effective for capturing the specific attributes necessary for assessing the value of scientific ideas.

\subsection*{Framework} The current design of our framework shows considerable promise in the automated assessment of scientific ideas, yet there are avenues for further enhancing the evaluator’s performance. The quantitative assessment of scientific ideas is inherently complex, involving a blend of objective metrics and subjective judgments. Our method, leveraging the representations of large language models, demonstrates the potential to approximate human judgment to a significant degree, which will provide human reviewers with an objective score, rather than replacing them to give subjective comments from multiple perspectives.

\begin{table*}[!htb]
\renewcommand{\arraystretch}{1.4}
\centering
\begin{tabular}{l|l}
\toprule[2pt]
\textbf{Title}    & \textbf{SMART: Self-supervised Multi-task pretrAining with contRol Transformers}                                                                                                                                                                                                                                                                                                                                                                                                                                                                                                          \\
\textbf{Abstract} & \begin{tabular}[c]{@{}l@{}}Self-supervised pretraining has been extensively studied in language and vision domains, \\ where a unified model can be easily adapted to various downstream tasks by \\ pretraining representations without explicit labels. When it comes to sequential \\ decision-making tasks, however, it is difficult to properly design such a pretraining \\ approach that can cope with both high-dimensional perceptual information and the \\ complexity of sequential control over long interaction horizons ...\end{tabular}                                    \\
\textbf{Scores}   & \textbf{Human-rated Score:} 7.50\qquad\qquad \textbf{LLM Representation Score:} \textcolor{blue}{\textbf{7.22}}                                                                                                                                                                                                                                                                                                                                                                                                                                                                                                                              \\ \midrule[1pt]
\textbf{Title}    & \textbf{Ensemble Homomorphic Encrypted Data Classification}                                                                                                                                                                                                                                                                                                                                                                                                                                                                                                                               \\
\textbf{Abstract} & \begin{tabular}[c]{@{}l@{}}Homomorphic encryption (HE) is encryption that permits users to perform computations \\ on encrypted data without first decrypting it. HE can be used for privacy-preserving \\ outsourced computation and analysis, allowing data to be encrypted and outsourced to \\ commercial cloud environments for processing while encrypted or sensitive data. \\ HE enables new services by removing privacy barriers inhibiting data sharing or\\ increasing the security of existing services ...\end{tabular}                                                    \\
\textbf{Scores}   & \textbf{Human-rated Score:} 1.50\qquad\qquad \textbf{LLM Representation Score:} \textcolor{blue}{\textbf{1.61}}                                                                                                                                                                                                                                                                                                                                                                                                                                                                                                                           \\ \midrule[1pt]
\textbf{Title}    & \textbf{Comparative Analysis between Vision Transformers and CNNs from Neuroscience}                                                                                                                                                                                                                                                                                                                                                                                                                                                                                          \\
\textbf{Abstract} & \begin{tabular}[c]{@{}l@{}}Neuroscience has provide many inspirations for the development of artificial intelligence, \\ especially for neural networks for computer vision tasks. Recent research on animals' \\ visual systems builds the connection between neural sparsity and animals' levels of \\ evolution, based on which comparisons between two most influential vision architecture, \\ Transformer and CNN, are carried out. In particular, the sparsity of attentions in \\ Transformers is comprehensively studied, and previous knowledge on sparsity of ...\end{tabular} \\
\textbf{Scores}   & \textbf{Human-rated Score:} 2.50\qquad \textbf{LLM Representation Score:} \textcolor{red}{\textbf{4.80 (Over Estimated)}}                                                                                                                                                                                                                                                                                                                                                                                                                                                                                                             \\ \midrule[1pt]
\textbf{Title}    & \textbf{Neural Causal Models for Counterfactual Identification and Estimation}                                                                                                                                                                                                                                                                                                                                                                                                                                                                                                            \\
\textbf{Abstract} & \begin{tabular}[c]{@{}l@{}}Evaluating hypothetical statements about how the world would be had a different course \\ of action been taken is arguably one key capability expected from modern AI systems. \\ Counterfactual reasoning underpins discussions in fairness, the determination of blame \\ and responsibility, credit assignment, and regret. In this paper, we study the evaluation \\ of counterfactual statements through neural models ...\end{tabular}                                                                                                                   \\
\textbf{Scores}   & \textbf{Human-rated Score:} 7.33\qquad \textbf{LLM Representation Score:} \textcolor{red}{\textbf{5.21 (Under Estimated)}}                                                                                                                                                                                                                                                                                                                                                                                                                                                                                                            \\ \bottomrule[2pt]
\end{tabular}
\caption{Case study of the comparision between human-rated scores and LLM representation scores}
\label{table_case_study}
\end{table*}

\clearpage

\begin{sidewaystable*}
\begin{center}
\small
\renewcommand{\arraystretch}{2.2}
\begin{tabular}{c|c|ccl|cc|cc}
\toprule[1pt]
\multirow{2}{*}{Domain}                       & \multirow{2}{*}{Num} & \multicolumn{3}{c|}{Mean}                                                                     & \multicolumn{2}{c|}{Min} & \multicolumn{2}{c}{Max} \\
                                            &                      & Human Mean                     & Ours Mean                      & Diff                        & Human Min   & Ours Min   & Human Max   & Ours Max  \\ \midrule
Deep Learning and representational learning & 527                  & 5.5131 $\pm$ 1.3967 & 5.3385 $\pm$ 1.0125 & 3.17\%                      & 1.6667      & 1.7936     & 8.0000      & 8.0428    \\
Social Aspects of Machine Learning          & 168                  & 5.4062 $\pm$ 1.4132 & 5.1867 $\pm$ 1.0791 & 4.06\%                      & 1.5000      & 1.0039     & 8.0000      & 7.2320    \\
Reinforcement Learning                      & 232                  & 5.4297 $\pm$ 1.5289 & 5.2762 $\pm$ 1.1101 & 2.83\%                      & 1.5000      & 2.2600     & 8.0000      & 8.2055    \\
General Machine Learning                    & 111                  & 5.4955 $\pm$ 1.2828 & 5.1640 $\pm$ 1.0825 & 6.03\%                      & 2.3333      & 1.7422     & 8.0000      & 7.4929    \\
Applications                                & 249                  & 5.6285 $\pm$ 1.4025 & 5.1934 $\pm$ 1.0932 & 7.73\%                      & 1.0000      & 2.3488     & 8.0000      & 7.6288    \\
Theory                                      & 86                   & 6.1155 $\pm$ 1.2607 & 5.3296 $\pm$ 1.0076 & 12.85\%                     & 2.0000      & 3.5655     & 8.0000      & 8.3156    \\
Generative models                           & 72                   & 5.6245 $\pm$ 1.3514 & 5.4084 $\pm$ 1.0636 & 3.84\%                      & 1.5000      & 1.5492     & 8.0000      & 7.7018    \\
Unsupervised and Self-supervised learning   & 95                   & 5.5672 $\pm$ 1.2967 & 5.2882 $\pm$ 0.9917 & 5.01\%                      & 2.0000      & 2.9334     & 8.0000      & 7.2551    \\
Optimization                                & 76                   & 5.3518 $\pm$ 1.3639 & 5.0908 $\pm$ 0.9983 & 4.88\%                      & 1.5000      & 1.5517     & 8.0000      & 7.2960    \\
Machine Learning for Sciences               & 96                   & 5.2392 $\pm$ 1.6701 & 5.0426 $\pm$ 1.1133 & \multicolumn{1}{c|}{3.75\%} & 1.6667      & 2.3116     & 8.0000      & 7.3444    \\
Neuroscience and Cognitive Science          & 23                   & 6.0667 $\pm$ 1.3181 & 5.1115 $\pm$ 0.9040 & 15.75\%                     & 3.0000      & 2.7473     & 8.0000      & 6.6993    \\
Probabilistic Methods                       & 49                   & 5.6442 $\pm$ 1.4135 & 5.2455 $\pm$ 0.9943 & 7.06\%                      & 2.5000      & 3.3245     & 8.0000      & 7.9918    \\
Infrastructure                              & 18                   & 5.9648 $\pm$ 1.2578 & 5.2065 $\pm$ 1.0139 & 12.71\%                     & 3.5000      & 3.0230     & 8.0000      & 7.1837    \\
None                                        & 5                    & 3.5500 $\pm$ 1.6763 & 3.6157 $\pm$ 1.8071 & 1.85\%                      & 1.0000      & 0.34500    & 5.7500      & 5.3299    \\ \bottomrule[1pt]
\end{tabular}
\end{center}
\caption{The detailed distribution of LLM representation scores and human-rated scores in 14 domains.}
\label{table:score_details}
\end{sidewaystable*}

\end{document}